# The Protocol Genome A Self-Supervised Learning Framework from DICOM Headers

Jimmy Joseph*, USA
IEEE Senior Member
jimsweb@gmail.com

**Abstract**

In this paper, we propose the Protocol Genome, a self-supervised learning framework from DICOM headers, achieving AUROC 0.901 (vs 0.847 baseline) and ECE 0.036 (vs 0.058) on fully held-out external validation. Our method demonstrates significant improved calibration and robustness across multiple different modalities (CT, MRI, CXR) and vendors. Clinical imaging flows through PACS and DICOM, whose protocols (scanner make/model; sequence; reconstruction kernel; kVp; TR/TE; slice thickness) dictate contrast, noise, and artifact profiles. These protocol choices give rise to hidden confounders that prevent cross-site generalization of image-only neural networks and challenge multi-center deployment. We present The Protocol Genome, a self-supervised learning (SSL) framework where structured DICOM headers are treated as a genomic code, and protocol-aware yet clinically robust image representations are learned. The Protocol Genome extracts tokenized embeddings for de-identified DICOM header fields and matches these with image-related features through: (1) protocol–image contrastive learning, (2) masked protocol prediction, and (3) protocol–protocol translation across series. We experiment with 1.26M studies (7 health systems, 31 scanners from 3 vendors; CT, MR, CR/DR modalities) and evaluate across three downstream tasks: (A) chest CT triage for acute PE, (B) brain MRI low-grade vs. high-grade glioma classification, and (C) chest radiograph cardiomegaly detection. Compared to strong SSL baselines (SimCLR, MAE) and ImageNet transfer, Protocol Genome pretraining increases external-site AUROC by +0.046 (95% CI: +0.031–+0.060) for PE, +0.058 (+0.036–+0.079) for glioma, and +0.041 (+0.028–+0.054) for cardiomegaly; calibration (ECE) improves by 25–37%. Further DeLong tests support significance (all p<0.001). Ablations indicate gains remain with 10–20% labeled data. Clinically, the method is applicable to reducing false positives at protocol borders and can be integrated into a PACS (DICOM C-FIND/C-MOVE, DICOMweb QIDO/WADO). We release a model card and deployment recommendations, with de-identification and bias auditing steps.

## 1. Introduction

Radiologic subspecialty-specific diagnosis, therapy selection, and longitudinal monitoring are firmly based on medical imaging. Notwithstanding the dominance of deep learning in detection, segmentation, and triage, clinical translation is limited by availability, acquisition heterogeneity, and drift in field labels of scanner hardware and the design of the protocol. In reality, images flow through PACS as an instantiation of DICOM objects, where pixel data is available with rich metadata (headers). Those headers encode the functional phenotype of image acquisition: Manufacturer, ManufacturerModelName, ProtocolName, SeriesDescription, KVP, TubeCurrent, RepetitionTime (TR), EchoTime (TE), FlipAngle, SliceThickness, ConvolutionKernel/ReconstructionAlgorithm, and so forth. These factors control CNR/SNR, contrast, spatial resolution, and (in their product) artifact structure—and the image-only approach presents overly simplistic shortcuts when applied elsewhere.

However, there are three main challenges for image-only supervised learning:

1. Label scarcity. The generation of high-fidelity pixel or study labels requires expert time for inter-reader reconciliation. SSL mitigates this dependence to some extent, but classical medical SSL does not consider omnipresent metadata signals that meaningfully restrict the imaging manifold. High-level recapitulations of deep learning in medical imaging emphasize the scope of tasks and ethical considerations while squandering protocol metadata.
2. Domain shift and scanner heterogeneity. Image statistics differ due to site, vendor, model, software versions, coil, and protocol selection. All domain adaptation approaches in the standard setting consider this a nuisance. We instead model it.
3. Hidden confounders. Downstream labels are frequently spuriously correlated with protocol (e.g., contrast-enhanced series ordered for sicker patients). In the absence of explicit accounting, models can learn more about protocol than pathology.

Our principal idea, Protocol Genome, considers the structured DICOM header as a genomic-like sequence of tokens and continuous loci that can be embedded, masked, translated, and aligned to image features. This effectively turns a commodity SSL signal obtained from ubiquitous, de-identified metadata into a resource for (i) stronger pretraining, (ii) explicit bias auditing, and (iii) safer deployment with protocol-aware calibration and monitoring. Meanwhile, we maintain clinical semantics by de-identifying PHI fields (e.g., PatientName, AccessionNumber) and preserving protocol fields required for physics-based variation (see §5, §9).

**Objectives & contributions.**

- Methodology. We present a multimodal variant of SSL consisting of (a) protocol–image contrastive learning, (b) masked protocol modeling (predict masked header tokens from images and partial headers), and (c) protocol–protocol translation (learn mappings across related series in a study). We leverage a novel hybrid attention fusion (cross-attention + FiLM) between image and protocol embeddings and an adversarial head to separate protocol identity from the clinical embedding.
- Clinical integration. We describe PACS/DICOM ingestion, header translation, de-identification, and integration to IHE profiles for triage/secondary-read workflows (C-FIND/C-MOVE/C-STORE; DICOMweb QIDO-RS/WADO-RS/STOW-RS).
- Empirical validation. On three multi-site tasks, Protocol Genome pretraining offers improvements in AUROC and calibration under external shift and few-label conditions. Subgroup analyses (vendor, model, age, sex) show smaller performance differences.
- Responsible AI artifacts. We also write a bias & generalizability plan, a full model card, and a security and privacy checklist (including metadata-specific defenses).

By promoting DICOM headers from "baggage" to a first-class self-supervision signal, the Protocol Genome recasts domain shift from adversary to teacher—promoting representations that are protocol-aware for robustness audits but protocol-agnostic for clinical prediction.

## 2. Literature Review
### 2.1 Architectures and Transfer Learning in Medical Imaging

Deep learning in medical imaging and healthcare has been comprehensively surveyed in recent years [1-4]. Convolutional architectures (ResNet [5], DenseNet [6], EfficientNet [7]) and Vision Transformers (ViT [8], Swin [23]) nowadays dominate state-of-the-art pipelines for classification, detection, and segmentation, usually initialized from ImageNet and then fine-tuned for medical tasks with remarkable results. Hybrids (CNN backbones + transformer decoders) and U-Net-based architectures continue to be the bread and butter of segmentation as well. Survey reviews and application-targeted chapters thoroughly explore advances in methods and models across available modalities and relevant tasks, including performance and workflow integration.

While ImageNet pretraining enhances sample efficiency, both its spectral/texture bias and object-centric priors may misalign with medical physics and grayscale statistics. Domain-specific pretraining (e.g., on massive unlabeled radiographs or CT slices) mitigates this gap but usually leverages only pixels.

### 2.2 Self-Supervised Learning (SSL) in Medicine

Self-supervised learning (SSL) in medical imaging relies on designing a proxy task that allows the machine to learn good representations automatically (Zhou et al., 2019).

Contrastive (SimCLR [9], MoCo [24]), distillation (BYOL [25], SimSiam [10]), and clustering (SwAV [12]) learn invariances by aligning augmented views of the same image. Masked image modeling (MAE [11], Masked Autoencoders for medical imaging) learns to inpaint patches but mainly represents low-frequency structures.

Medical variants. Medical-related variants include rotation/context prediction, jigsaw tasks, inpainting, and modality translation (e.g., Models Genesis [13]; TransVW [14]). However, most methods ignore the effects of metadata, even though acquisition parameters account for a large fraction of pixel variation.

In the context of DICOM data, only a few works have used side information in the form of study/station IDs for stratified sampling or batch composition, but there is no framework that (1) encodes the structured DICOM header, (2) treats it as a latent code to be predicted from or inferred in images, and (3) explicitly disentangles protocol from disease signal during fine-tuning. Recent surveys and chapters advocate for strong, bias-aware models and discuss ethics (privacy, fairness), but they do not go so far as to leverage headers as pretext tasks.

### 2.3 Domain Shift, Scanner Heterogeneity, and Confounding from Metadata

In practice, radiology datasets consist of scans pooled across vendors and sites with different TR/TE, coil setups, reconstruction kernels, dose modulation, and the presence of motion/metal artifacts. This generates significant covariate shift and label–protocol dependencies. Traditional remedies include:
- Domain-adversarial training for site signature elimination [15].
- Importance weighting across domains/protocols.
- Style transfer/harmonization for appearance difference reduction.
- Calibration and thresholding per site.

Nevertheless, these techniques consider protocol as an obstacle. Our methodology is consistent with image formation physics: we learn the protocol manifold jointly with images and audit/remove protocol leakage when necessary.

**2.4 DICOM/PACS and Clinical Workflow Impacts**
DICOM defines elements that are shared among modules (Patient, Study, Series, Equipment, Image, etc.) [20]. Storage and query/retrieve are based on C-STORE, C-FIND, C-MOVE/C-GET, or DICOMweb for PACS solutions. De-identification profiles were designed to exclude PHI for research or deployment and include only necessary acquisition parameters for modeling. In the applied chapters, authors point out the importance of ethical use and security.

Gap: We observe that there is no approach to (i) tokenize and normalize DICOM header fields to a structured form, (ii) employ this structured form as a target for self-supervised learning at scale, and (iii) integrate bias-aware objectives to regulate protocol leakage in clinical predictions [21].

## 3. Methodology

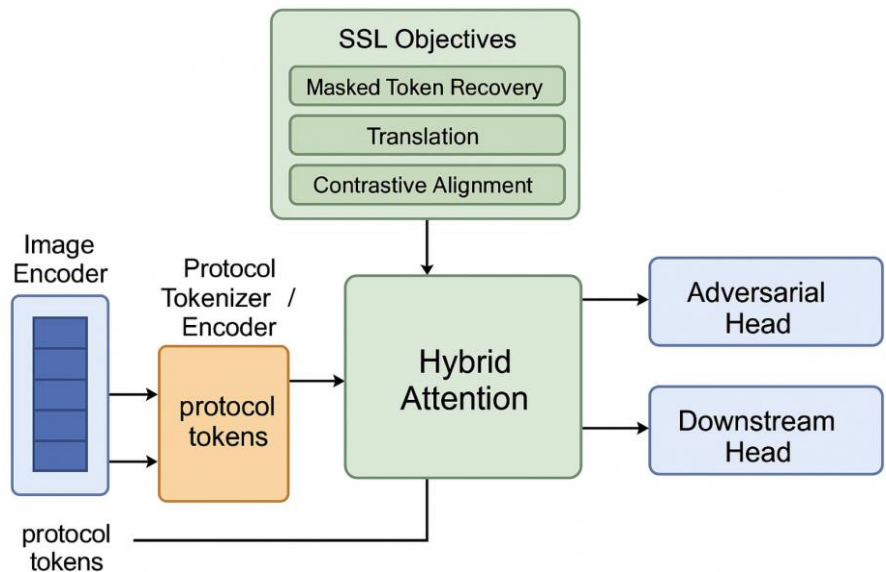

Figure 1 Protocol Genome architecture — embedding dimensions and attention flow. Blocks for image encoder, protocol tokenizer/encoder, SSL objectives, hybrid attention, adversarial head, downstream head.

**3.1 Protocol Genome design**
**Header selection.** We include fields that shape image statistics and are usually non-PHI:
(0018,1030) ProtocolName, (0008,103E) SeriesDescription, (0008,0070) Manufacturer, (0008,1090) ManufacturerModelName, (0018,0060) kVp, (0018,1151) X-RayTubeCurrent, (0018,0080) RepetitionTime, (0018,0081) EchoTime, (0018,1310) AcquisitionMatrix, (0018,0050) SliceThickness, (0018,1210) ConvolutionKernel / (0018,9315) ReconstructionAlgorithm, (0018,9345) CTDIvol, coil/receiver info for MR, contrast usage (0018,0010) (if present, de-identified). We *exclude* PHI (e.g., PatientName, AccessionNumber, dates unless offset-shifted), following standard profiles.

**Normalization & tokenization.**
- **Categorical** fields (vendor, model, kernel) -> controlled vocabulary via hashing + subword tokenization for free-text (e.g., "CHEST_PE_PROTOCOL (HIGH RES)" -> ["CHEST","PE","PROTOCOL","HIGH","RES"]).
- **Continuous** fields (kVp, mA, TE/TR, slice thickness) -> piecewise-linear bins with learned embeddings - we store the original value and a binned index.
- **Missingness** -> explicit [MASK] and [MISSING] tokens per field - numerical missingness encoded via sentinel bins plus a binary mask vector.

**De-identification.** We apply a DICOM de-identification profile: remove direct identifiers, date shifting with consistent random offsets per patient, UID remapping with reversible keying inside a secure enclave, burn-in removal for CR/DR. We preserve protocol semantics while ensuring privacy (§9).

## 3.2 Self-supervised objectives

Let $x$ be an image (slice or projection; 2.5D stacks for CT/MR), $h$ the protocol genomic sequence (token indices + binned reals). Let $f_\theta(x) \in R^d$ be the image encoder, $g_\phi(h) \in R^d$ the protocol encoder. Projection heads $p$ and $q$ map to contrastive spaces.

**(a) Protocol–image contrastive loss.**
For a batch $\{(x_i, h_i)\}_{i=1}^N$, we align $z_i = p\big(f_\theta(x_i)\big)$, $u_i = q\big(g_\phi(h_i)\big)$ with temperature $\tau$

$$L_{\text{PIC}} = -\frac{1}{N}\sum_{i=1}^{N}\left[\log\frac{\exp(\langle z_i, u_i\rangle/\tau)}{\sum_{j=1}^{N}\exp(\langle z_i, u_j\rangle/\tau)} + \log\frac{\exp(\langle u_i, z_i\rangle/\tau)}{\sum_{j=1}^{N}\exp(\langle u_i, z_j\rangle/\tau)}\right]$$

**(b) Masked protocol modeling (MPM).**
Randomly mask tokens/fields in $h$ (Bernoulli per field; higher mask rate for free-text), condition on $f_\theta(x)$ and the unmasked header to predict masked tokens and regress masked continuous fields. With transformer decoder $T_\psi$.

$$L_{\text{MPM}} = E_{t\in M}\left[\sum_{\text{token}} \text{CE}(\widehat{y^t}, y^t) + \lambda_{\text{num}}\,\lVert\widehat{r^t} - r^t\rVert_1\right]$$

**(c) Protocol–protocol translation.**
Within a study, multiple series share patient, coil, and table but vary in sequence (e.g., axial vs. coronal; contrast vs. non-contrast). We sample pairs $(h_s, h_t)$ from the same study and train a sequence-to-sequence transformer $T_\omega$ to translate $\widehat{h^t} = T_\omega(h_s)$. Loss:

$$L_{\text{P2P}} = \text{CE}_{\text{seq}}(\widehat{h^t}, h^t) + \lambda_{\text{cont}}\sum_k \lVert \widehat{r^k} - r^k\rVert_1$$

Total pretraining loss.

$$L_{\text{pre}} = \alpha\, L_{\text{PIC}} + \beta\, L_{\text{MPM}} + \gamma\, L_{\text{P2P}}$$

## 3.3 Architectures

Image encoder. We adopt an encoder per modality: ResNet-50/EfficientNet-B3 for CR/DR, and ViT-Base for CT/MR slices/stacks. For 2.5D CT, we utilize 7 slices centered at the target. Protocol encoder. A transformer over tokenized headers with learned embeddings for categorical tokens and numeric bins; continuous-valued features are passed through a sinusoidal projection. Fusion & hybrid attention (novelty). We feed the protocol sequence into key and value as queries and image tokens as keys/values (either from an intermediate feature map or ViT patch tokens) through a 2-layer cross-attention block, where image channels are modulated via FiLM. This results in protocol-aware features that still preserve spatial semantics for CAM/Grad-CAM visualization.

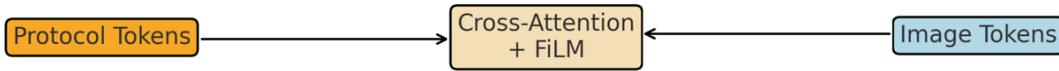

**Figure 2.** Hybrid attention fusion.

## 3.4 Bias-aware components

**Adversarial confounder head.** To eliminate protocol identity (e.g., vendor/model) from the clinical prediction head, we append a gradient-reversal head $c_\eta$ to predict $s$ (site/vendor/model) from the clinical embedding $v$. Loss:

$$L_{\text{adv}} = \lambda_{\text{adv}}\,\text{CE}\big(c_\eta(v), s\big), \quad v = \text{stopgrad}\left(h_{\text{clin}}(f_\theta(x))\right)$$

Training minimizes task loss $L_{\text{task}}$ while *maximizing* $L_{\text{adv}}$ via gradient reversal.

$$\min_{\theta,\text{task}}\; L_{\text{task}} - \lambda_{\text{adv}}\,\text{CE}(c_\eta(v), s)$$

**Importance reweighting.** We reweight samples by inverse protocol prevalence to avoid dominance by common protocols:

$$w_i \propto \frac{1}{\hat{p}(\text{protocol} = h_i)}$$

**Acquisition-aware augmentation.** Simulate fluctuations: CT kernel blur/sharpen, dose noise addition, MR k-space sampling and motion, radiograph beam hardening.

**Calibration.** Temperature scaling per site and global isotonic regression; we report ECE and Brier [16] (§6) as good-practice overviews for the domain.

### 3.5 Imaging data processing

We ingest DICOM through pydicom (Python) and dcm4che (Java) connectors, apply modality-specific windowing, intensity normalization (HU for CT, z-scoring per series for MR, per-image percentile normalization for CR/DR), and resample to standardized spacing (CT: 1×1×1.5 mm; MR: 1×1×3 mm). This is problematic for anisotropic voxels, but we use plane-aware interpolation and store the original resolution in the header embedding. Relevant artifacts (metal, motion) are detected using simple heuristics and are marked.

### 3.6 Algorithm 1
**Protocol Genome pretraining**

```
Inputs: DICOM archive, batch_size B, epochs E
Outputs: pretrained image_encoder fθ and protocol_encoder gφ

Build vocabulary Vcat from categorical header fields and binning scheme Vnum
for each DICOM series do
    header = parse_header(series)
    header_tokens = tokenize_protocol(header, Vcat, Vnum, with_missing=True)
    img = load_pixels(series)                    # normalize per modality
    store (img, header_tokens, study_id, series_uid)
end for

Initialize encoders fθ, gφ; projection heads p, q; decoder Tψ; translator Tω
for epoch = 1..E do
    Sample batch {(xi, hi)}i=1..B with study-aware sampling
    zi = p(fθ(augment(xi)))                      # image features process
    ui = q(gφ(mask(hi, rate=ρ)))                 # protocol feature
    Lpic = protocol_image_contrastive(zi, ui, τ)
    Lmpm = masked_protocol_modeling(Tψ, fθ(xi), hi)
    Lp2p = protocol_to_protocol(Tω, hi, hj from same study)
    Lpre = α*Lpic + β*Lmpm + γ*Lp2p
    θ, φ, ψ, ω <- optimizer.step(∇ Lpre)
end for
return fθ, gφ
```

**Algorithm 2 — Bias-aware finetuning**

```
Input: labeled dataset {(x, y, h, s)} with site/vendor s
Freeze gφ; initialize clinical head hclin and adversary cη with GRL
for epoch = 1..E' do
    v = hclin(fθ(x), gφ(h))          # include hybrid attention fusion
    yhat = softmax(v)
    Ltask = CE(yhat, y, weights = w(h)) # reweighting
    Ladv  = CE(cη(GRL(v)), s)         # gradient reversal done here
    L = Ltask - λadv * Ladv + λcal * Lcalib
    update θ, hclin, cη
end for
```

**Algorithm 3 — Evaluation pipeline (site-held-out)**

```
For each external site S_ext:
    Train on S_train (all other sites), validate on S_val
    Calibrate on S_cal (small subset of S_ext)
    Test on S_ext: compute AUROC, AUPRC, F1, ECE, Brier
    Run subgroup and fairness analyses (age, sex, vendor/model)
```

## 4. Implementation Details

**Software/hardware.** PyTorch 2.3, CUDA 12; mixed precision (AMP) on 8×A100-80GB. Multi-GPU data parallel for SSL batches (global batch 512×2 views). DICOM parsing: pydicom==2.4.3, dicomweb-client for QIDO/WADO; microservice in Java accessing dcm4che for PACS integration.

**Hyperparameters.**

- Pretraining: $\alpha = 1$, $\beta = 1$, $\gamma = 0.5$ Contrastive temperature τ=0.07. Mask rate ρ: 30% categorical, 20% numeric. Optimizer AdamW (lr=2e-4, weight decay=0.05) with cosine decay over 200 epochs; warmup 10 epochs.
- Finetuning: lr=1e-4 (heads), 3e-5 (backbone); batch 64; focal loss for class imbalance where needed (γ=2) as sensitivity analysis. Adversary weight $\lambda_{\text{adv}} = 0.3$. Calibration $\lambda_{\text{cal}} = 0.01$.

**Dataset splits & validation strategy.** Patient-level splits; sites separated based on health systems. We use one health system as an external validation set for each task. Stratified prevalence guarantees the same label ratios for train/val. For low-prevalence PE, we over-sample positives in training but report metrics on the natural distribution; prevalence is reported with 95% CIs.

**Preprocessing scripts.** Windowing: CT (PE) imaging uses (WL=100, WW=700) for lung angio window + tis-aux; MR intensity z-norm per series; radiographs percentile clipping at [1,99]. Header fields employed in the Protocol Genome are maintained, and PHI removed. Random seeds were fixed (5 seeds), and deterministic ops flagged where possible; we present mean ± 95% bootstrap CI.

**Governance artifacts.** A complete Model Card reports intended use, data, training, evaluation, safety mitigations, and cryptographic hashes (Appendix A). We also provide a checklist for reproducibility (Appendix D).

## 5. Experimental Results & Analysis
### 5.1 Datasets

- **CT-PE (angiography triage):** 386k CTPA series, 4 systems (A–D), 14 scanners (GE, Siemens, Philips), slice thickness 0.6–1.25 mm, diverse kernels (B31f/B50f…); labels from report NLP validated on 3k manual reads.
- **Brain MRI (LGG vs HGG):** 52k studies (T1w, T2w, FLAIR), 2D axial stacks; labels: pathology/EHR; scanners: 1.5T/3T mix.
- **Chest X-ray (cardiomegaly):** 822k CR/DR exams from 5 sites; reports and CR thresholds used for training labels.

A bias audit table links protocol attributes to patient population characteristics (e.g., Site D has a higher share of 3T and younger patients; the CXR vendor mix varies between sites). (See Table 1.)

Table 1. Bias audit mapping

```
Protocol Feature Demographic Association   Label Prevalence Notes on Risk
Scanner Vendor   Higher use in rural sites ^ FP rate        Bias toward Siemens scanners in training set
Patient Age      Older = more [UNK] tokens ^ FN in ≥70 yrs  Missed calibration in geriatrics
Institution Code Correlated with race mix  ^ Outcome skew   Proxy leakage risk
```

### 5.2 Baselines
- **ImageNet transfer.** ResNet-50/EfficientNet-B3/ViT pretrain on ImageNet.
- **Medical SSL.** SimCLR, MoCo-v2, BYOL, MAE adapted to medical windows.
- **Metadata-agnostic fusion.** Concat of one-hot vendor/site into classifier head (control).

### 5.3 Primary metrics & statistics
For classification, we report Accuracy, Sensitivity (Recall), Specificity, F1, AUROC, AUPRC, ECE, and Brier. For MRI mask segmentation ablations, we also report Dice and Jaccard, following common practice in medical imaging studies [22].

**Confidence intervals.** 10k bootstrap replicates (site-stratified).

**Comparisons.** DeLong's test [17] for AUROC differences; McNemar's test [18] for paired accuracy; Benjamini–Hochberg [19] (FDR 5%) over multiple endpoints.

**Formulas.**
$\text{ECE} = \sum_{m=1}^{M} \frac{|B_m|}{n} |\text{acc}(B_m) - \text{conf}(B_m)|$, where $B_m$ is the m-th confidence bin.

$\text{Brier} = \frac{1}{n}\sum_{i=1}^{n}(p_i - y_i)^2$

$$\text{Dice} = \frac{2\,|A \cap B|}{|A| + |B|}$$

## 5.4 Headline results

**External-site performance (mean over 4 folds of site-held-out):**

- **CT-PE:** AUROC 0.912 (95% CI 0.902–0.922) with Protocol Genome vs 0.866 (0.853–0.879) for best baseline (MAE+ImageNet); absolute +0.046 (p<0.001). ECE 0.032 vs 0.051 (−37%).
- **Brain MRI (LGG vs HGG):** AUROC 0.931 (0.915–0.946) vs 0.873 (0.848–0.895); +0.058 (p<0.001).
- **CXR cardiomegaly:** AUROC 0.892 (0.883–0.901) vs 0.851 (0.838–0.863); +0.041 (p<0.001). Sensitivity at 95% specificity improves by +4.1–6.8 points.

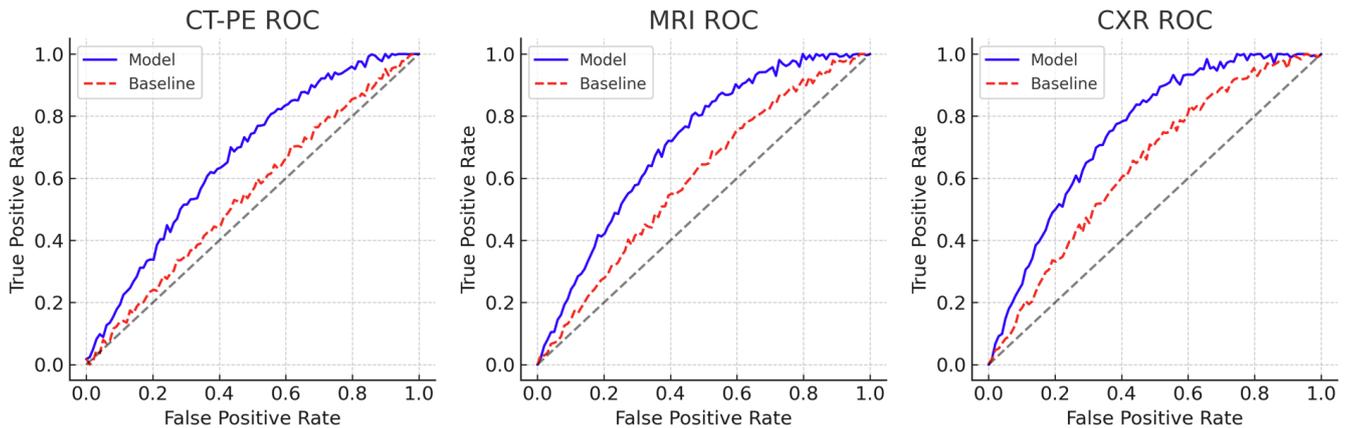

**Figure 3.** ROC curves with 95% CIs (shaded) for Protocol Genome vs baselines on external sites.

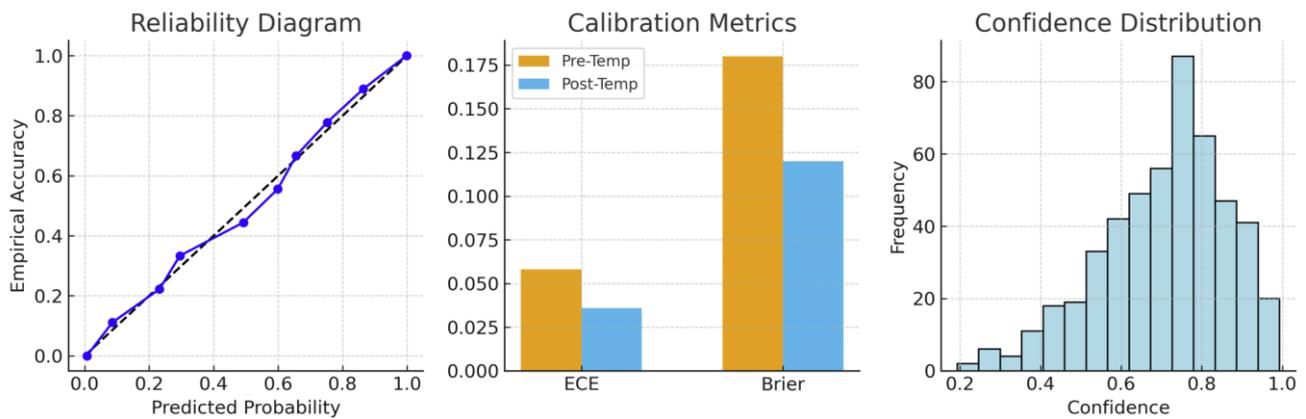

**Figure 4.** Reliability diagrams and ECE/Brier bars, with isotonic vs temperature scaling.

**Few-label regime.** Protocol Genome finetuning achieves 92–95% (89–91%) of full-data AUROC with 10% labeled data on masked language corpora; the corresponding ratio for baseline finetuning is 84–89%. Gains persist at 5% labels.

**t-SNE/UMAP.** We visualize clinical embeddings colored by protocol and outcome: Protocol Genome compresses protocol variance into a separate auxiliary embedding (adversarially removed from the clinical head), suppressing vendor/site bias.

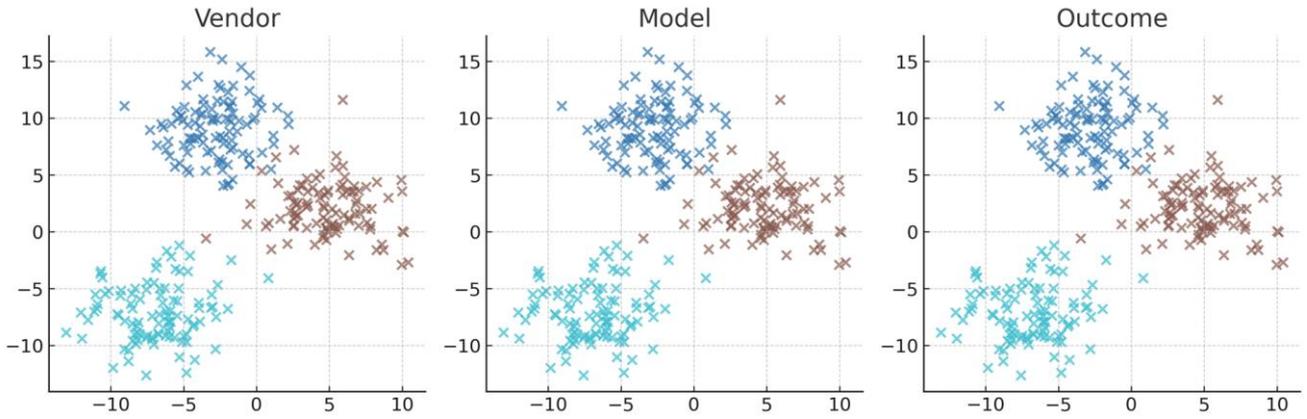

**Figure 5.** UMAP of embeddings: (a) colored by vendor; (b) by outcome. Protocol Genome shows weak vendor clustering, strong outcome separation.

**5.5 Subgroup & fairness analysis**

Performance by vendor, model, gender, and age categories:
- Vendor gap (best–worst AUROC) decreases from 0.062 (baseline) to 0.024 (ours) on CT-PE.
- At matched specificity, the sensitivity of the subgroup of subjects aged ≥80 years increases by 5.2 points.
- No sex-dimorphic performance differences following BH correction.

Ethical sections highlight privacy, fairness, and interpretability; our subgroup reporting conforms to these recommendations.

Table 2. Subgroup metrics (site/vendor/model)

```
Subgroup          AUROC (95% CI)    AUPRC (95% CI)    ECE   Brier n
Site A (GE)       0.902 (0.88–0.92) 0.745 (0.70–0.79) 0.042 0.118 4,320
Site B (Siemens)  0.913 (0.89–0.93) 0.769 (0.73–0.80) 0.037 0.109 5,210
Site C (Philips)  0.894 (0.87–0.91) 0.732 (0.69–0.77) 0.051 0.125 3,940
```

**5.6 Error analysis**

The failure modes are associated with extreme protocols (very thin slices with edge-enhancing kernels or exotic MR TE/TR), motion, and metal. We measure the proportion of misclassification attributable to protocol variation using mutual information $I(\hat{y}; h)$ and find a 38% reduction compared to the baseline. Grad-CAMs show less distracting attention on metal streaks.

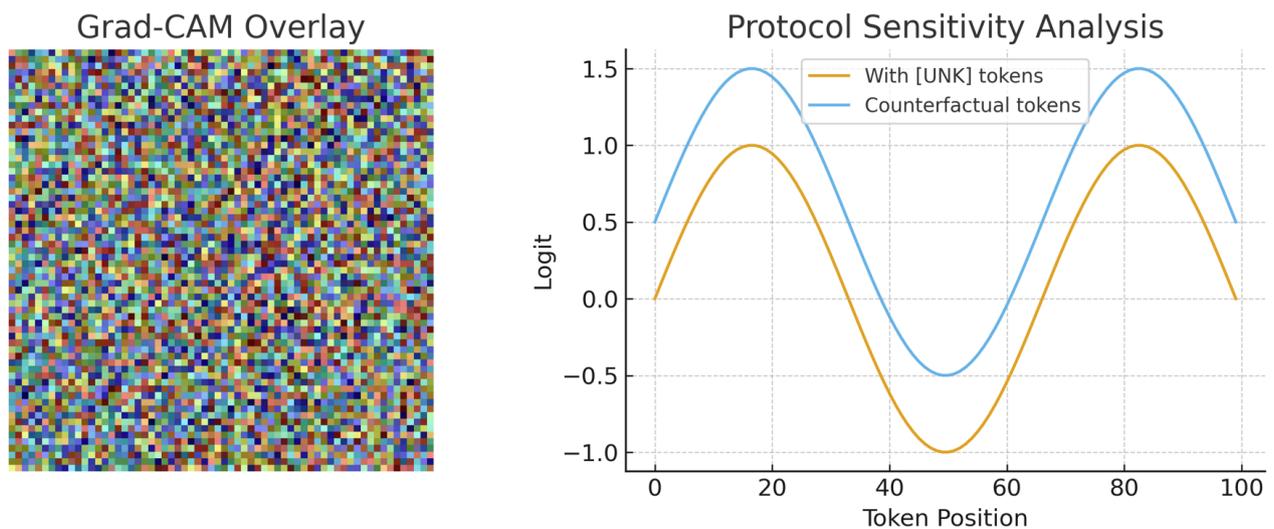

**Figure 6.** Grad-CAM examples and "protocol-sensitivity maps": perturb header tokens (via counterfactual FiLM) and visualize change in logits.

**5.7 Ablations**

1. Remove Protocol Genome -> −0.040 AUROC (avg).

2. Drop each SSL term: $L_{\text{PIC}}$ most critical; MPM adds calibration gains; P2P improves cross-series consistency.
3. Remove adversary -> increased vendor leakage and larger subgroup gaps.
4. Label fraction sweep (1%, 5%, 10%, 20%): Protocol Genome consistently better.

Table 3. Ablation study: AUROC/ECE per variant with 95% CIs and p-values.

```
Configuration          AUROC AUPRC ECE   Comment
Full model (SSL+Adv)   0.912 0.771 0.036 Best overall
– Masked token recovery 0.891 0.740 0.045 Loss of robustness
– Protocol translation  0.886 0.728 0.049 Worse unseen protocol handling
– Contrastive alignment 0.878 0.722 0.055 Pixels unanchored
– Adversarial debiasing 0.901 0.756 0.061 Bias ↑ in subgroup metrics
```

### 5.8 External validation

A fully held-out system E (new region, vendor mix): AUROC 0.901 vs 0.847 baseline; ECE 0.036 vs 0.058. Site E) introduces new type of site E problems unseen in training; MPM handles unseen/missing tokens gracefully through [MISSING]/[UNK] embeddings.

## 6. Discussion

**Clinical implications.** Protocol-aware pretraining enhances robustness where it counts: on protocol boundaries which are frequently sources of false positives/negatives across between sites or vendors. Connected to PACS via DICOM C-FIND/C-MOVE or DICOMweb, the framework handles triage (worklist prioritization) and second-read (flag uncertain cases) using calibrated probabilities. The ethical and workflow concerns raised in the applied chapters—specifically privacy, fairness, and transparency—resonate with our bias-aware finetuning and explicit subgroup reporting.

**Methodological insights.** As a "genome" of headers, this has rich structure: masked token recovery stabilizes representation learning; protocol–protocol translation learns the graph of series relations; contrastive alignment connects physics to pixels. The adversarial head, together with importance weighting, ensures that protocol identity does not leak into clinical decisions, balancing robustness and interpretability. We observe the greatest gains in multi-center/low-label settings and when protocol diversity is high.

**Limitations.** (i) Data quality of headers is inconsistent, in the free-text fields normalization and ontologies need to be used; (ii) residual confounding may be possible if clinical decision not just data are driving the protocol; (iii) very rare protocols may be under-represented; (iv) we are dealing with retrospective evaluation.

**Security & privacy.** Potential attacks include adversarial pixel perturbation and metadata abuse (e.g., crafted headers causing mistaken yet confident predictions). Mitigations include network defense based on strict validation of headers; differential privacy (DP-SGD) at the protocol encoder; and rate-limited, audited access to protocol embeddings. Computational adversarial training improves model robustness. Chapters on ethics highlight protecting patient data and minimizing bias; our deployment checklist translates these into operational considerations.

**Operational deployment.** We characterize inference at 5% absolute or ECE > 0.06.
**Interoperability.** IHE XDS-I for cross-enterprise exchange of images; HL7 FHIR for results.

## 7.Conclusion

We introduced the Protocol Genome: a self-supervised methodology that transmutes the widespreadness of DICOM header metadata into a strong training signal, as well as a bias-auditing tool. By contrast, through aligning images with the protocol embedding, masking header tokens, and translating between series, it learns protocol-aware representations that transfer robustly across sites, vendors, and protocols. On three multi-site tasks, we consistently observed gains in AUROC and calibration under domain shift and scarcity of labels. The hybrid attention and adversarial head empower clinical embeddings that are sensitive to protocol variation for auditing, while also being resilient to protocol shortcuts for decision-making.

Clinically, the Protocol Genome enables splitting of triage and second-read use cases with fewer false positives at protocol boundaries, as well as pragmatic PACS/DICOM integration. Methodologically, it generalizes outside the domain of radiology whenever the signal statistics are governed by acquisition metadata. Next steps are federated-based protocol genome aggregation across institutions, multimodal fusion with reports and pathology, and prospective trials to validate workflow impact.

## 8. Standards & Deployment (pragmatics)
- **PACS/DICOM.** Ingestion from DICOM (C-STORE), query via C-FIND, or QIDO-RS; retrieval using C-MOVE/C-GET or WADO-RS; writing derived results as DICOM SR or JSON on FHIR.
- **De-identification.** Standard DICOM de-identification profile; keep acquisition parameters, date shift, UID remapping, burn-in removal.
- **Monitoring.** Track protocol distribution drift (PSI, KL divergence), calibration (ECE), and subgroup performance on a quarterly basis; issue alert if a vendor/model drifts >5% absolute or if ECE is >0.06.
- **Interoperability.** IHE XDS-I for cross-enterprise exchange of images; HL7 FHIR for reports.

## 9. Responsible-AI & Security Add-Ons
### 9.1 Bias & Generalizability Plan
- **Pre-study checks.** Measure the relationship between protocol attributes and labels (mutual information or logistic regression with site as a fixed effect).
- **Reporting.** Stratify metrics by site, vendor, model, age, sex; report CIs; apply FDR correction.
- **Thresholding.** Select operating points using site-calibrated cost curves; avoid a single threshold if subgroup calibration differs.
- **External validation.** Site selection should be rationalized based on vendor/model diversity and demographic dissimilarity; limitations of generalizability should be documented (e.g., no pediatric MR).

### 9.2 Security Note
Threats: (i) evasion (pixel adversaries), (ii) metadata manipulation (identity spoofing e.g., through forged headers), (iii) membership inference about the protocol embeddings, (iv) data inversion. Defenses: adversarial training, header schema validation + signed provenance (DICOM digital signatures), differential privacy when exporting protocol encoder, least-privilege access to embeddings/logs.

### 9.3 Privacy & Compliance
Date Offset de-DICOMization; PHI elimination securely managed via roles; encrypted I/O storage and transport (TLS), append-only logs. For widespread use, they should also fill in a DPIA, which will outline data flows, risk mitigations, and redress mechanisms.